\documentclass[letterpaper, 10 pt, conference]{ieeeconf}

\IEEEoverridecommandlockouts
\usepackage{amsmath}
\usepackage{amsfonts}
\usepackage{mathtools}
\usepackage{hyperref}
\hypersetup{hidelinks}
\usepackage{booktabs}
\usepackage{xcolor-material}
\usepackage{multirow}
\usepackage{capt-of}
\usepackage{amssymb}
\usepackage{xcolor}
\usepackage{standalone}
\usepackage{cite}
\usepackage[nameinlink,capitalise,noabbrev]{cleveref}

\title{\LARGE \bf
What Would Removing This Corruption Change? \\ Failure Attribution for Vision-Language-Action Models
}

\author{
Carson Sobolewski, Anjali Parashar, Fei Chen, Yingke Li, and Chuchu Fan
\thanks{The authors are with the Massachusetts Institute of Technology, USA.
        {\tt\small \{csobo, anjalip, feic, yingkeli, chuchu\}@mit.edu}.}
}
\newcommand{\baselinename}[1]{\textcolor{MaterialPurple900}{\textsf{#1}}}

\begin{document}
\IEEEaftertitletext{
    \centering
    \includegraphics[width=0.94\textwidth]{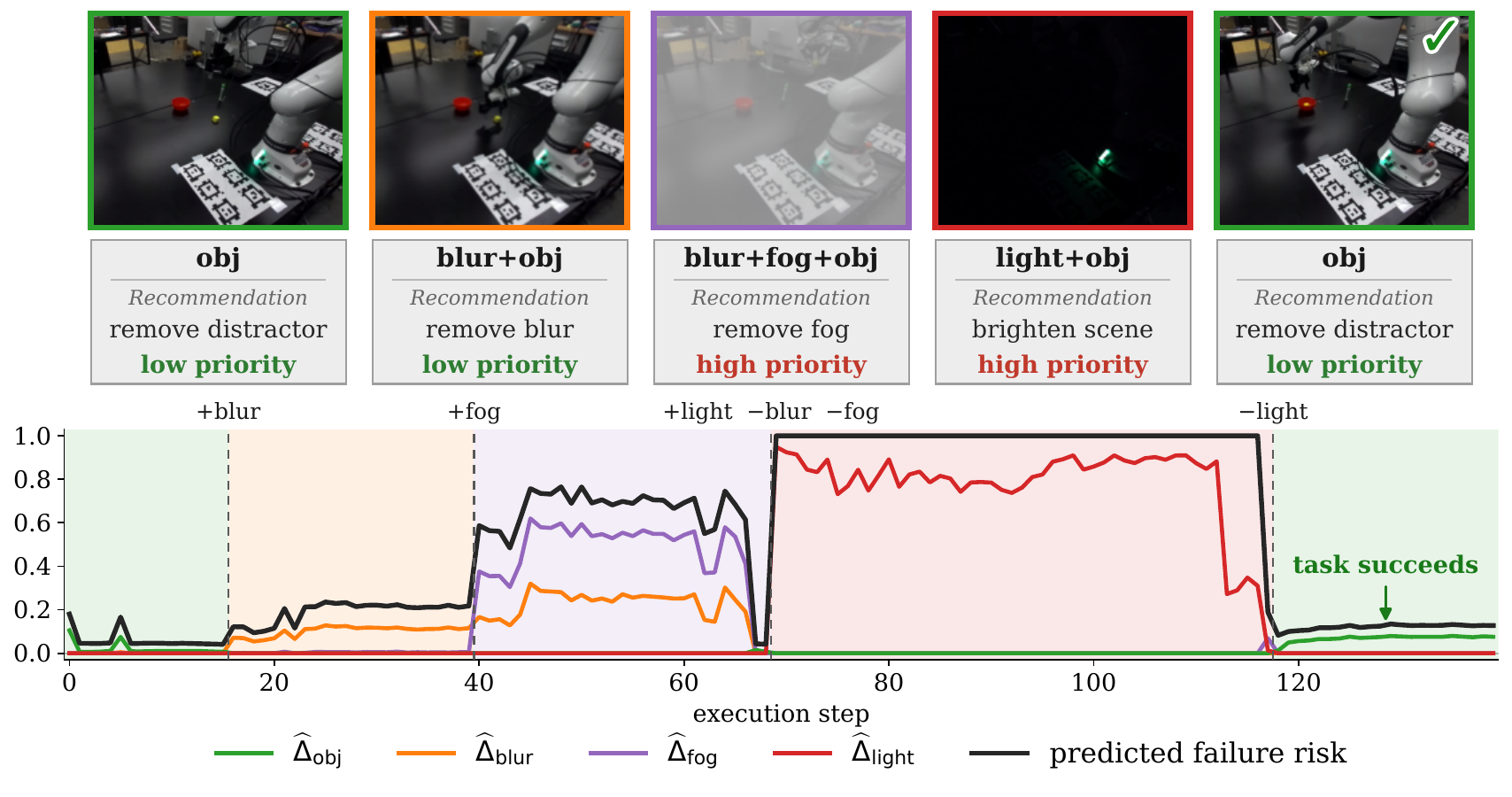}
    \captionof{figure}{Example online failure monitoring and per-corruption attribution during a $\pi_{0.5}$-DROID hardware rollout. Corruptions are introduced in sequence throughout a rollout (shaded regions): a distractor object, then blur, then fog alongside blur, then near-darkness replacing fog and blur. The plot shows the predicted failure risk $\hat r$ and per-corruption attribution estimates $\hat\Delta_k$ at every execution step. Because of their high $\hat\Delta_k$ values, fog and light receive high-priority correction recommendations, while the distractor receives low attribution. After lighting is corrected, predicted risk falls and the task succeeds.}
    \label{fig:hardware_rollout}
    \vspace{1em}
}
\maketitle
\thispagestyle{empty}
\pagestyle{empty}

\begin{abstract}
Vision-language-action (VLA) models are sensitive to visual corruptions, but knowing that a corruption is typically harmful does not establish whether removing it would benefit the current episode. Furthermore, when several corruptions co-occur, deciding what to correct requires understanding which ones contribute to failure risk and how much their removal would help. We define this problem as \emph{failure attribution}. Detecting corruption presence alone cannot answer this question, as corruptions can be present but relatively harmless, while directly measuring removal benefits requires costly paired policy rollouts with and without the corruption present. To address this, we introduce a factorized risk model that learns separate representations of the scene and corruptions, allowing us to estimate removal benefits through latent intervention from the current image and task instruction, without paired clean images or additional counterfactual rollouts. We evaluate our approach on OpenVLA and $\pi_{0.5}$ in simulation and $\pi_{0.5}$-DROID on a Franka robot. Our method outperforms monolithic and vision-language-model (VLM) baselines, including achieving approximately 3x lower attribution error than frontier VLMs. We also demonstrate our method adapts to execution context and corruption intensity, and can be used for online failure monitoring on hardware.
\end{abstract}

\section{Introduction}

Vision-Language-Action (VLA) models~\cite{kim2024openvla,black2024pi0,black2025pi05} have emerged as generalist robotic policies capable of performing manipulation tasks directly from images and natural language instructions. However, their reliance on visual input makes them sensitive to visual corruptions~\cite{fei2026liberoplus}. For example, viewpoint changes and sensor noise have considerable impact on OpenVLA's performance, reducing success rate from 76.5\% to 1.1\% and 19.3\%, respectively~\cite{fei2026liberoplus}. These effects motivate knowing how each corruption influences a policy during execution. A straightforward approach is to estimate average failure rates for each corruption combination based on historical data. However, these averages assign the same prediction to episodes with the same corruption combination, regardless of corruption intensity or the current scene context. Knowing that a corruption is harmful on average therefore does not establish whether removing it would reduce failure risk in the current episode.

When several corruptions co-occur, this problem is compounded: we need to determine which corruptions contribute to the current failure risk and which would offer the greatest benefit if removed. We call this \emph{failure attribution}: quantifying each corruption's contribution by asking, \emph{``How much would failure risk decrease if this corruption were removed?''}
Answering this question can guide whether and how to correct an execution failure.

However, failure attribution presents two challenges.
First, corruption presence alone does not establish its effect on performance. A corruption may be clearly visible yet have little impact on a particular policy, so detecting it does not indicate whether correcting it would help.
Prior works address neighboring problems, including improving robustness through corruption-inclusive training~\cite{fei2026liberoplus,xie2026strongvla} or input restoration~\cite{orjuela2026improving}, and predicting failure through online monitoring~\cite{gu2025safe,li2026sentinelvla,agia2024sentinel,xu2025faildetect}. However, these approaches do not directly answer the attribution question of how much removing each corruption would reduce failure risk.

Second, directly measuring this removal benefit requires intervention: executing the policy repeatedly with and without each corruption from the same underlying state. Because a corruption's effect depends on which others remain present, exhaustively characterizing these interactions requires covering a large majority of combinations, rather than just single corruptions. Such comparisons are expensive in simulation and difficult to reproduce during deployment, where repeated failed executions can also put hardware at risk. We therefore seek to estimate removal effects from the available observation, without executing these counterfactual rollouts.

To address these challenges, we introduce a factorized risk model that estimates each corruption's contribution to failure risk from the current image and task instruction alone.
The model learns separate latent representations of the scene and corruptions. By intervening on the corruption latent while holding the scene latent fixed, it estimates how failure risk would change if a corruption were removed (\cref{fig:method}), without modifying the physical scene or rerunning the policy.

We evaluate the model on OpenVLA and $\pi_{0.5}$ in simulation and $\pi_{0.5}$-DROID on hardware. Our experiments validate that our approach accurately estimates per-corruption attributions, captures interactions between co-occurring corruptions, and learns sensitivity to scene context and corruption intensity. Under co-occurring corruptions, it achieves approximately 3x lower attribution error than frontier VLMs. Hardware experiments demonstrate our approach can be used for online monitoring (\cref{fig:hardware_rollout}) as corruptions change during an episode. In summary, we:
\begin{enumerate}
\item Formulate failure attribution for VLA models as estimating the reduction in failure risk from removing individual visual corruptions.
\item Introduce a factorized risk model that estimates these removal effects through latent interventions, using only the current image and task instruction.
\item Evaluate the method on OpenVLA and $\pi_{0.5}$ in simulation and $\pi_{0.5}$-DROID on hardware, demonstrating attribution under co-occurring corruptions, sensitivity to execution context and corruption intensity, and online failure monitoring.
\end{enumerate}

\section{Related Work}

\paragraph{Robustness Analysis}
Prior works evaluate VLA robustness to visual and environmental disturbances through benchmarks such as LIBERO-Plus~\cite{fei2026liberoplus} and The Colosseum~\cite{pumacay2024colosseum}, as well as adversarial attacks~\cite{wang2025exploringadversarial}. Other approaches improve robustness through corruption-aware training~\cite{fei2026liberoplus,xie2026strongvla} or input restoration~\cite{orjuela2026improving}. These works characterize or mitigate aggregate sensitivity, but do not attribute an observed failure to the responsible corruption, which is the gap this work addresses.

\paragraph{Failure Detection and Uncertainty Quantification}
Online failure monitors use internal policy representations~\cite{gu2025safe,li2026sentinelvla}, action consistency~\cite{agia2024sentinel}, failure-free training~\cite{xu2025faildetect}, out-of-distribution detection~\cite{sinha2022ood}, or conformal prediction~\cite{ren2023knowno}. VLM-based methods provide post-hoc failure explanations~\cite{liu2023reflect,duan2025aha}, while image-region attribution links reliance on nuisance features to poor performance~\cite{zhang2026embodied}. These approaches predict \emph{whether} failure is likely or identify generic visual cues, but do not quantify the counterfactual effect of removing a specific corruption. Our factorized latent model estimates this effect without paired counterfactual trajectories.

\section{Problem Statement}
\label{sec:problem}

\begin{figure}[t!]
    \centering
    \includegraphics[width=\columnwidth]{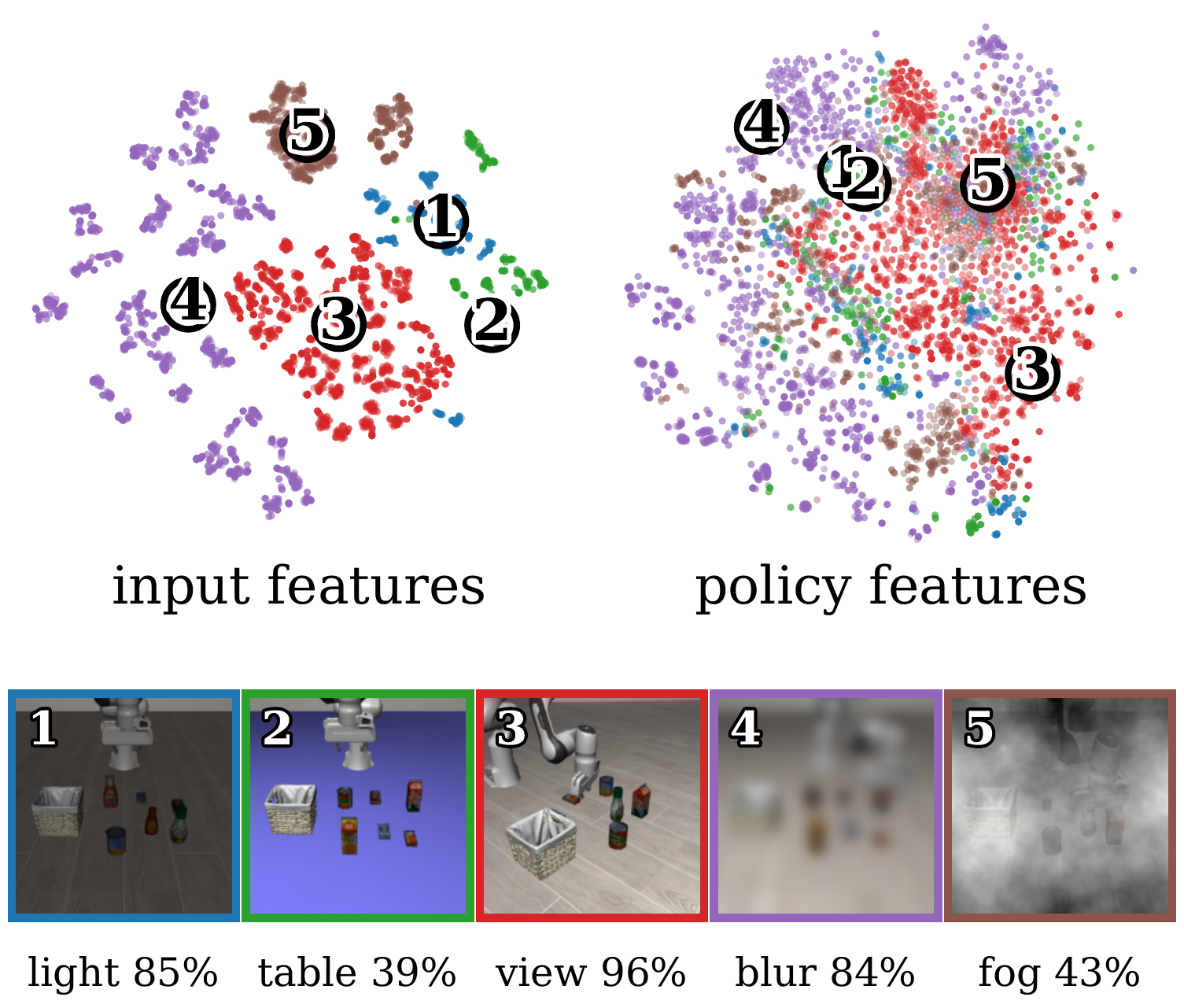}

    \caption{
        t-SNE projections of input features (left) and internal policy features (right) for the same held-out initial observations from failed OpenVLA episodes. Colors denote corruption types, and numbered markers indicate the location of one example image per corruption (shown below, with empirical failure rates). Corruption types are more visibly separated in the input features than in the internal policy features.
    }
    \label{fig:feature_tsne}
\end{figure}
We consider a VLA policy that is rolled out for an episode of maximum length $T$. At step $t\in\{0,\dots,T-1\}$, the policy receives observation $o_t=(\phi_t,q_t)$ consisting of a camera image $\phi_t$ and robot state $q_t$, along with an episode-wide task instruction $\ell$. The episode outcome provides a binary failure indicator $y\in\{0,1\}$, denoting whether the policy failed to complete the task within the allotted $T$ steps, with $y=1$ indicating failure.

We consider $K$ known visual corruption types, and let $c_t\in\{0,1\}^K$ indicate which are present in $\phi_t$. We assume access to a dataset of $N$ labeled episodes for this information, $\mathcal{D}=\left\{\left(\{\phi_t^{(i)}\}_{t=0}^{T_i-1},\ell^{(i)},c^{(i)},y^{(i)}\right)\right\}_{i=1}^{N}$. In this dataset, corruptions remain fixed within each episode, so we use $c_t^{(i)}=c^{(i)}$. We model the observed image as the corrupted version of a clean image $\phi_t=\mathcal{T}(\phi_t^{\mathrm{clean}};c_t)$, where $\phi_t^{\mathrm{clean}}$ is an unobserved clean image. Let $s_t\coloneqq(\phi_t^{\mathrm{clean}},\ell)$ denote the corruption-free visual scene and task instruction. We define \emph{risk} as $r(c_t;s_t)\coloneqq P(y=1\mid c_t,s_t)$, the probability of failure given this context and the corruption vector.

Our objective is to identify the present corruptions and estimate how much failure risk would decrease if each were removed while holding the scene, instruction, and remaining corruptions fixed. Let $c_t^{\setminus k}$ denote $c_t$ with its $k$-th entry set to zero. Then, we define the attribution to corruption $k$ as the reduction in failure risk from removing it:
\begin{equation}
\label{eq:rf}
\Delta_k(c_t;s_t)
    =
        r(c_t;s_t)-r(c_t^{\setminus k};s_t)
      .
\end{equation}
Larger \(\Delta_k\) indicates greater removal benefit, with \(\Delta_k\approx r(c_t;s_t)\) meaning that removing corruption \(k\) nearly eliminates the risk. Estimating \(\Delta_k\) is non-trivial, as it requires comparing risks with and without \(k\) while holding the scene fixed. However, deployment provides neither the clean scene nor corruption labels. We mitigate the need for paired counterfactual episodes using latent representations for corruptions and predicting \(r(c_t;s_t)\), as we discuss next.

\section{Methodology}
\label{sec:method}
\begin{figure}
    \centering
    \includegraphics[width=0.9\linewidth]{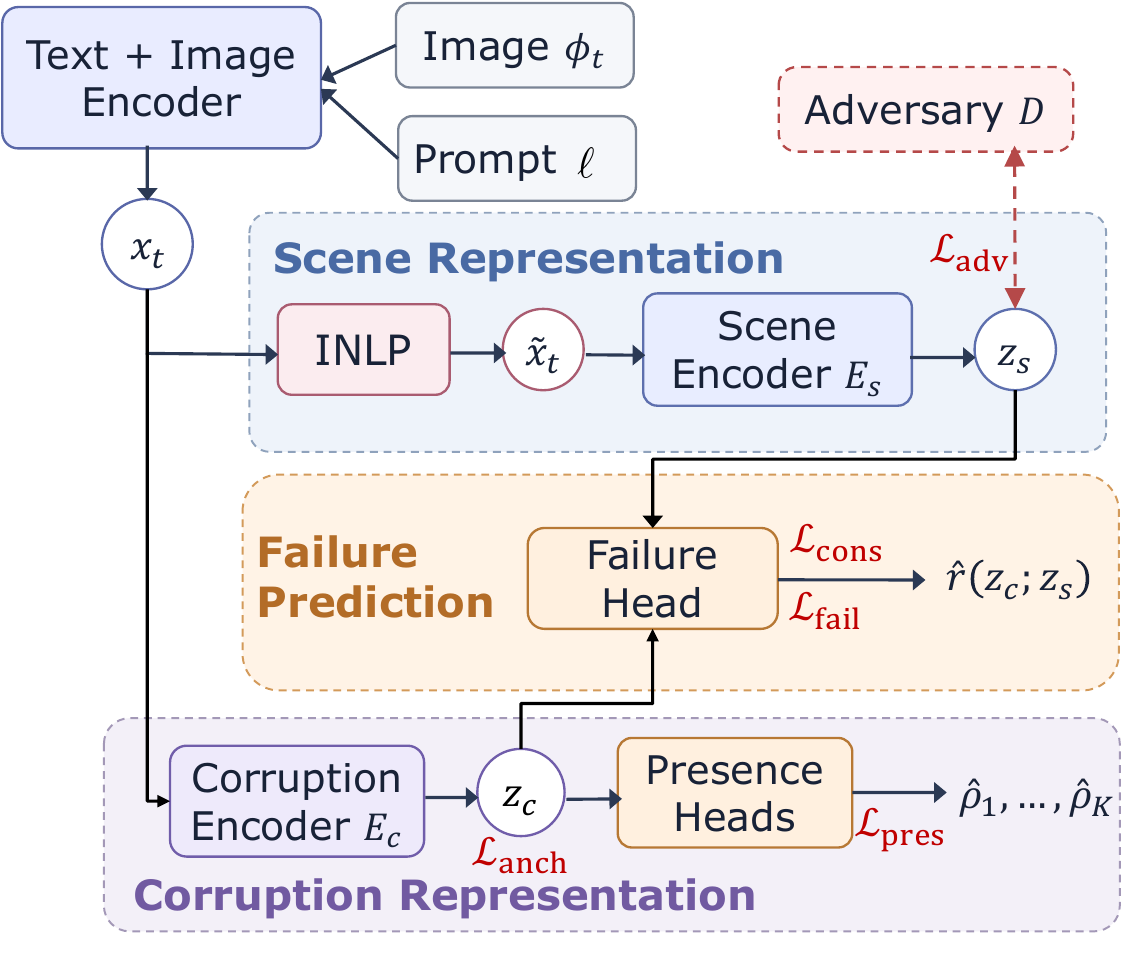}
    \caption{Factorized risk model for failure attribution.
    A frozen image-and-text encoder produces features
    used to construct scene and corruption latents,
    $z_s$ and $z_c$.
    INLP and adversarial training discourage corruption
    information in the scene branch, while presence and
    anchor losses structure the corruption latent.
    A shared failure head predicts risk from both latents.
    Attribution is estimated by zeroing the corresponding
    corruption block and reevaluating risk
    (\cref{eq:estrf}).
    Red labels indicate training objectives.}
    \label{fig:method}
\end{figure}

To estimate attributions without paired counterfactuals, we learn separate representations of the scene ($z_s$) and corruptions ($z_c$), standing in for $s_t$ and $c_t$, together with a failure predictor $\hat r(z_c;z_s)$ that approximates $r(c_t;s_t)$.

We approximate removal of corruption $k$ through a latent intervention: zeroing the corresponding portion of $z_c$, yielding $z_c^{\setminus k}$. Holding $z_s$ fixed, we estimate the per-corruption attribution as
\begin{equation}
\label{eq:estrf}
    \hat{\Delta}_k(\phi_t,\ell)
    = \operatorname{ReLU}\!\left(
        \hat r(z_c;z_s)
        - \hat r(z_c^{\setminus k};z_s)
      \right),
\end{equation}
where the ReLU restricts the reported attribution to positive risk reductions. This replaces a comparison requiring an unobserved clean scene and an explicit corruption vector with a latent intervention. For this estimate to be meaningful, the representations must support interpreting a zeroed portion of $z_c$ as corruption absence while preserving the underlying scene information in $z_s$. We next describe how these representations are constructed (\cref{sec:representations}) and how our training objectives encourage this alignment (\cref{sec:disentanglement}).
\cref{fig:method} summarizes all the main components of our pipeline.

\subsection{Encoding Scene and Corruption Information}
\label{sec:representations}
A frozen text and image encoder maps the image and task instruction to features $x_t=f(\phi_t,\ell)\in\mathbb{R}^D$. Separate encoders construct the scene and corruption representations. The scene encoder $E_s$ receives projected features $\tilde x_t$, obtained from a fixed linear projection designed to remove linearly decodable corruption information (\cref{sec:disentanglement}), and produces $z_s=E_s(\tilde x_t)\in\mathbb{R}^{d_s}$. A corruption encoder $E_c$ takes the original features $x_t$ and outputs $z_c=E_c(x_t)\in\mathbb{R}^{Kd_c}$. We use SigLIP encoder~\cite{zhai2023siglip} for our experimental analysis but also report an ablation on an alternative model-based encoder in \cref{sec:safe}.

We partition $z_c$ into $K$ blocks, one per corruption type, $z_c=(z_c^{(1)},\dots,z_c^{(K)})$, with $z_c^{(k)}\in\mathbb{R}^{d_c}$. Each block has a linear presence head that estimates the corresponding corruption's presence probability $\hat\rho_k(\phi_t,\ell)= \sigma\!\big(w_k^\top z_c^{(k)}+b_k\big)$. Although corruption labels remain fixed within each training episode, presence is predicted independently for each frame and can therefore vary during execution, as demonstrated in \cref{fig:hardware_rollout}.
A shared failure head takes both representations and outputs the estimated risk $\hat r(z_c;z_s)$.

The factorized model with separate encoders for scene and corruption, and compartmentalization of $z_c$ by corruption type alleviates the need for pairwise counterfactual rollouts, and allows us to easily estimate \cref{eq:estrf}. The latent intervention in \cref{eq:estrf} sets $z_c^{(k)}$ to zero while leaving all other blocks and $z_s$ unchanged. All $K$ attribution estimates reuse the same representations, requiring only one encoding pass and $K+1$ failure-head evaluations.

\subsection{Training Objectives for Disentangled Representations}
\label{sec:disentanglement}

The architecture alone does not ensure that latent interventions correspond to corruption removal. Each corruption block must encode its designated corruption, with $z_c^{(k)}=0$ representing its absence. Meanwhile, corruption information retained in $z_s$ could allow the failure head to bypass $z_c$, weakening or suppressing the estimated attributions. We therefore combine objectives that enforce the structure of $z_c$, reduce corruption leakage into $z_s$, and train risk predictions before and after latent intervention.

Using the dataset $\mathcal D$, we jointly optimize the scene encoder $E_s$, corruption encoder $E_c$, presence heads, and failure head with the objectives below (also shown in \cref{fig:method}). The feature encoder and fitted projection remain fixed.

\paragraph{Presence loss} A binary cross-entropy loss between the estimated presence probabilities and presence labels
\begin{equation}\label{eq:l1}
    \mathcal{L}_{\mathrm{pres}} = \frac{1}{K}\sum_{k=1}^{K}\mathrm{BCE}\big(\hat\rho_k,\,c_{t,k}\big),
\end{equation}
where $\mathrm{BCE}(p,q)\coloneqq-\,q\log p-(1-q)\log(1-p)$, binds block $k$
to corruption $k$, so that $z_c$ carries the information needed to recover
the present corruptions.

\paragraph{Anchor loss} An anchor loss enforces that each block $z_c^{(k)}$ goes to zero when corruption $k$ is absent,
\begin{equation}\label{eq:l2}
    \mathcal{L}_{\mathrm{anch}} = \frac{1}{K}\sum_{k=1}^{K}\mathbb{E}\!\left[\,\big\|z_c^{(k)}\big\|_2^{2}\;\middle|\;c_{t,k}=0\,\right],
\end{equation}
where the expectation is the average over training samples with $c_{t,k}=0$. This loss encourages zero to represent corruption absence, providing the reference value used by the latent intervention.

\begin{figure}[t]
    \centering
    \includegraphics[width=\linewidth]
        {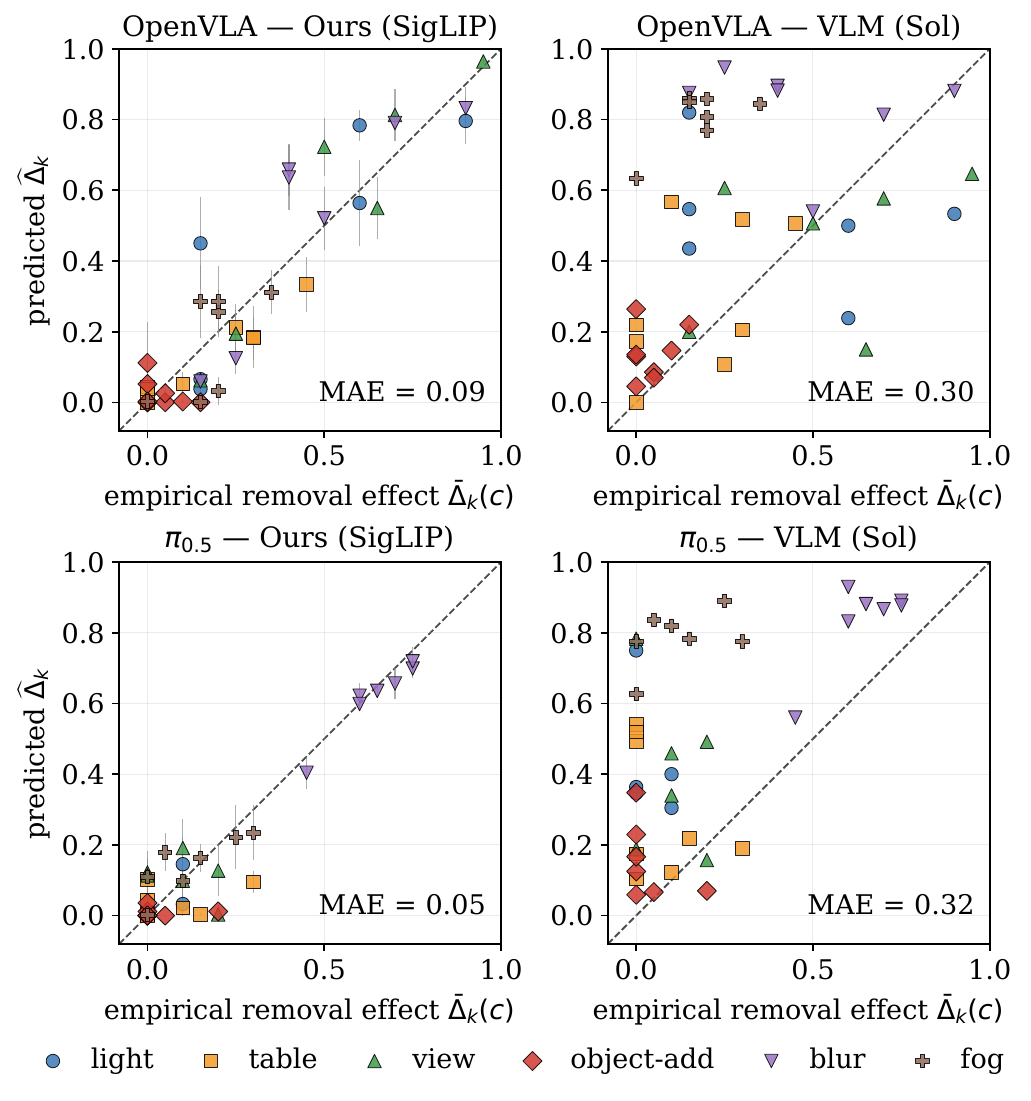}

    \caption{
        Predicted vs.\ empirical per-corruption attribution for OpenVLA (top) and $\pi_{0.5}$ (bottom). Columns compare \baselinename{Ours (SigLIP)} with
    rescaled responsibility scores from \baselinename{VLM (Sol)}. Each point is one present corruption within one corruption condition (e.g., the condition light+view contributes one point for light and one for view). Our estimates show closer agreement (diagonal), while the VLM's scores are less correlated with empirical per-corruption attribution.
    }
    \label{fig:calibration_grid}
\end{figure}

\paragraph{Nullspace projection} To remove linearly decodable corruption information from the scene branch, we fit an iterative null-space projection (INLP)~\cite{ravfogel2020inlp} on the frozen SigLIP features $x_t$. Linear probes predicting $c_t$ from $x_t$ are trained and projected out iteratively until corruption information is no longer linearly decodable. The projected features $\tilde x_t$ are passed to $E_s$, while $x_t$ is directly passed to $E_c$.

\begin{table}[t]
\centering
\small
\setlength{\tabcolsep}{6pt}
\renewcommand{\arraystretch}{1.12}
\caption{Failure attribution on LIBERO (2{,}500 episodes per policy). Attribution MAE (single-/multi-corruption conditions) and regret (multi-corruption failure episodes). Five-seed mean $\pm$ standard deviation. \emph{Oracle} achieves $0.000$ for all columns. \textbf{Bold}: best among non-oracle methods.}
\label{tab:main_attribution}
\resizebox{\columnwidth}{!}{%
\begin{tabular}{llccc}
\toprule
Policy & Method & Single\,$\downarrow$ & Multi\,$\downarrow$ & Regret\,$\downarrow$ \\
\midrule
\multirow{6}{*}{OpenVLA}
 & Monolithic     & $0.416_{\pm.000}$ & $0.265_{\pm.000}$ & $0.256_{\pm.025}$ \\
 & VLM (Terra)    & $0.207$ & $0.237$ & $0.285$ \\
 & VLM (Sol)      & $0.246$ & $0.308$ & $0.098$ \\
 & VLM (Opus\,5)  & $0.245$ & $0.292$ & $0.107$ \\
 & Ours (SAFE)    & $0.104_{\pm.012}$ & $0.117_{\pm.020}$ & $0.029_{\pm.010}$ \\
 & Ours (SigLIP)  & $\mathbf{0.095_{\pm.025}}$ & $\mathbf{0.101_{\pm.016}}$ & $\mathbf{0.011_{\pm.001}}$ \\
\midrule
\multirow{6}{*}{$\pi_{0.5}$}
 & Monolithic     & $0.141_{\pm.000}$ & $0.169_{\pm.000}$ & $0.521_{\pm.002}$ \\
 & VLM (Terra)    & $0.332$ & $0.252$ & $0.085$ \\
 & VLM (Sol)      & $0.405$ & $0.310$ & $0.026$ \\
 & VLM (Opus\,5)  & $0.336$ & $0.307$ & $0.026$ \\
 & Ours (SAFE)    & $0.048_{\pm.005}$ & $0.075_{\pm.007}$ & $0.028_{\pm.007}$ \\
 & Ours (SigLIP)  & $\mathbf{0.019_{\pm.006}}$ & $\mathbf{0.064_{\pm.006}}$ & $\mathbf{0.005_{\pm.004}}$ \\
\bottomrule
\end{tabular}%
}
\end{table}
\paragraph{Adversarial loss} To further discourage $z_s$ from carrying information about $c_t$, we perform adversarial training. Let $b_t=\mathbf{1}\{\sum_{k=1}^{K}c_{t,k}>0\}$ indicate whether any corruption is present. An adversarial discriminator $D$ (a small MLP) is trained to predict $b_t$ from $z_s$ by minimizing $\mathcal{L}_D=\mathrm{BCE}\big(D(z_s),\,b_t\big)$. The scene encoder is then trained against $D$ with a confusion loss that drives its prediction to chance,
\begin{equation}\label{eq:l3}
\mathcal{L}_{\mathrm{adv}}=\mathrm{BCE}\big(D(z_s),\tfrac{1}{2}\big).
\end{equation}
The uniform target encourages the discriminator's output toward $\frac{1}{2}$. Together with INLP, this objective discourages the scene representation from retaining corruption information that could bypass the intervention on $z_c$.

\paragraph{Failure loss} We jointly train the failure head and both encoders with a binary cross-entropy loss between the estimated risk and episode outcome, applied at every frame with the episode-wide label $y$:
\begin{equation}\label{eq:l4}
    \mathcal{L}_{\mathrm{fail}} = \mathrm{BCE}\big(\hat r(z_c;z_s),\,y\big).
\end{equation}
This loss is back-propagated to train both encoders and the failure head jointly.

\paragraph{Counterfactual consistency loss}
We encourage post-intervention risk predictions to agree, on average, with predictions for observations in which the corresponding corruption is absent.
For each corruption configuration $c$ represented in the minibatch and each $k$ with $c_k=1$, we compare the mean risk after intervening on corruption block $k$ with the mean risk under $c^{\setminus k}$, provided that both combinations occur in the batch:
\begin{equation}
\label{eq:l5}
\mathcal{L}_{\mathrm{cons}}=\frac{1}{|\mathcal P|}\sum_{(c,k)\in\mathcal P}\left(\bar r_c^{\setminus k}-\operatorname{sg}\!\left[\bar r_{c^{\setminus k}}\right]\right)^2.
\end{equation}
Here, $\mathcal{P}$ is the set of valid combination-removal pairs $(c,k)$ described above. Then, $\bar r_c^{\setminus k}$ is the mean post-intervention risk for observations with corruption vector $c$, and $\bar r_{c^{\setminus k}}$ is the mean risk without latent intervention for observations with corruption vector $c^{\setminus k}$. Both means are computed within the minibatch, and $\operatorname{sg}$ stops gradients through the target.

The full training objective is then
\begin{equation}
\begin{aligned}
\mathcal{L}
&= \mathcal{L}_{\mathrm{fail}}
 + \lambda_{\mathrm{pres}}\mathcal{L}_{\mathrm{pres}}
 + \lambda_{\mathrm{anch}}\mathcal{L}_{\mathrm{anch}} \\
&\quad + \lambda_{\mathrm{adv}}\mathcal{L}_{\mathrm{adv}}
 + \lambda_{\mathrm{cons}}\mathcal{L}_{\mathrm{cons}}.
\end{aligned}
\end{equation}
The discriminator $D$ is updated separately using $\mathcal{L}_D$ in alternation with the encoder and prediction-head updates. Together, these objectives train failure prediction and encourage the representation structure needed for latent interventions to approximate corruption removal. We evaluate their individual contributions in \cref{sec:latentfact}.

\section{Experiments}\label{sec:experiments}

\begin{figure}[t]
    \centering
    \includegraphics[width=\linewidth]
        {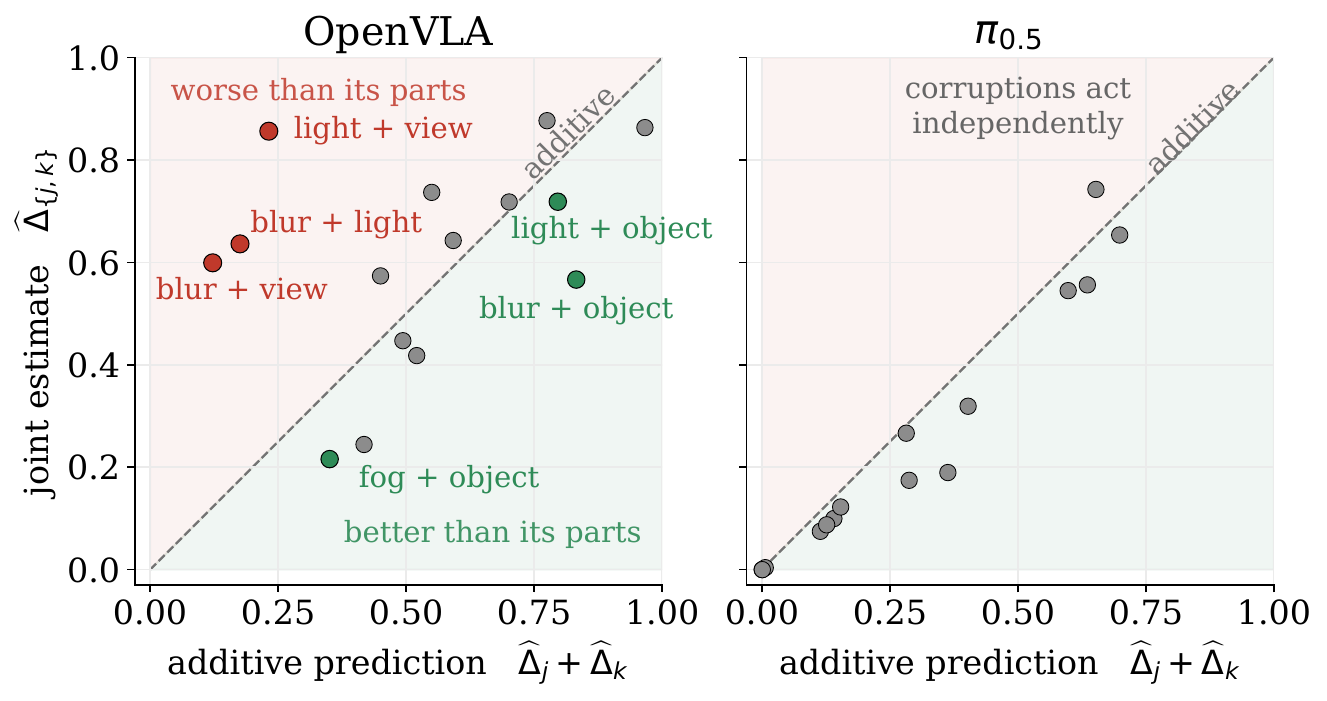}

    \caption{
        Joint versus summed individual attributions for two-corruption combinations under \baselinename{Ours (SigLIP)}. The vertical axis shows attribution from removing both corruptions together $\hat\Delta_{\{j,k\}}$. The horizontal axis sums their individual attributions. The diagonal marks perfect additivity. Several OpenVLA combinations lie far above the diagonal, indicating greater predicted benefit from joint removal than from the sum of individual removals. The $\pi_{0.5}$ estimates lie closer to the diagonal.
    }
    \label{fig:interaction}
\end{figure}
The central hypothesis of this paper is that attributing VLA failures to visual corruptions requires separating corruption from scene information, and that this factorization generalizes across policies, co-occurring corruptions, and hardware. We evaluate OpenVLA~\cite{kim2024openvla} and \(\pi_{0.5}\)~\cite{black2025pi05} in simulation in the LIBERO~\cite{liu2023libero} environment and pick-and-place tasks on a Franka robot using a DROID-fine-tuned \(\pi_{0.5}\) policy~\cite{khazatsky2024droid}. We compare against a monolithic presence-and-failure predictor, post-hoc VLM attribution, and a variant using internal policy features (SAFE~\cite{gu2025safe}), and ablate the training objectives. We discuss the key results in \cref{sec:results}, where we investigate whether corruption detection alone is sufficient for failure attribution, contribution of various components of our approach, and whether our attribution scheme captures co-occurring corruptions and learns different intensities of the same corruption. We also analyze how well the method transfers to hardware and online monitoring.

\subsection{Experimental Setup}

\subsubsection{Simulation}
We evaluate on the 10 LIBERO-Object tasks~\cite{liu2023libero} (\cref{fig:setup}, left) with six corruption categories from LIBERO-Plus~\cite{fei2026liberoplus}: light, table, view, object-add, Gaussian blur, and fog. We ran experiments on nominal images, all six individual corruptions, and 17 co-occurrence conditions: all 15 pairwise combinations, as well as two three-way combinations (fog and Gaussian blur, each with object-add and table). Episodes run for at most 280 steps. For each VLA, the dataset contains 2{,}500 episodes with 200 nominal and 100 per single- or multi-corruption condition.

\subsubsection{Hardware}
\begin{figure}[t]
    \centering
    \includegraphics[width=\linewidth]
        {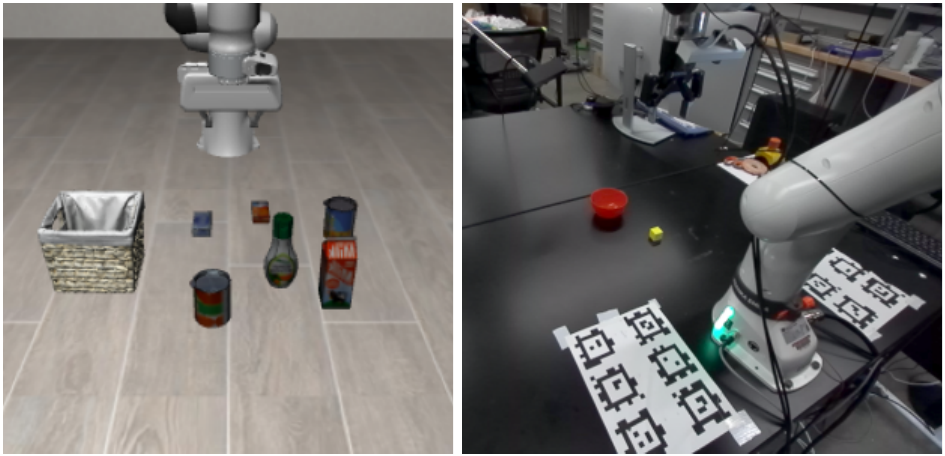}

    \caption{Example observations from the evaluation environments:
    LIBERO-Object simulation (left) and the Franka Panda hardware
    setup for the task ``put the cube in the bowl'' (right).
    We evaluate OpenVLA and $\pi_{0.5}$ in simulation
    and $\pi_{0.5}$-DROID on hardware.}
    \label{fig:setup}
\end{figure}

We evaluate the \(\pi_{0.5}\)-DROID policy~\cite{black2025pi05} on the DROID hardware platform~\cite{khazatsky2024droid} (\cref{fig:setup}, right) on a Franka Emika Panda with a Robotiq 2F-85 gripper\footnote{Additional video results for hardware are available at \url{https://vla-attribute.github.io}}. The task is ``put the cube in the bowl'' under four corruptions: physically reduced lighting and a distractor object, and synthetic Gaussian blur and fog applied to both camera streams. We test nominal and individual-corruption conditions, plus co-occurring blur and fog. Episodes are manually terminated and labeled as failures when the policy ceases making progress or risks collision. The dataset contains 150 episodes, with 25 per condition. The attribution model is trained solely on hardware episodes, whose corruption intensities differ from simulation.

\begin{table}[t]
\centering
\small
\setlength{\tabcolsep}{6pt}
\caption{Single-frame failure prediction ROC-AUC at different episode positions on held-out OpenVLA episodes (\baselinename{Ours (SigLIP)}, five-seed mean $\pm$ standard deviation). The oracle assigns test-set empirical corruption failure rates and is independent of episode position.}
\label{tab:failrecog}
\begin{tabular}{lcc}
\toprule
Episode position & Nominal & Corrupted \\
\midrule
0.0 & $0.510_{\pm.149}$ & $0.918_{\pm.017}$ \\
0.2 & $0.655_{\pm.093}$ & $0.918_{\pm.017}$ \\
0.4 & $0.485_{\pm.193}$ & $0.917_{\pm.017}$ \\
0.6 & $0.722_{\pm.062}$ & $0.932_{\pm.014}$ \\
0.8 & $0.679_{\pm.160}$ & $0.927_{\pm.017}$ \\
1.0 & $0.723_{\pm.162}$ & $0.925_{\pm.018}$ \\
\midrule
\emph{Oracle} & $0.500$ & $0.862$ \\
\bottomrule
\end{tabular}
\end{table}

\subsubsection{Evaluation Procedure}
Within our datasets, we perform a 60/20/20 train/validation/test split on the episode level. Each episode is downsampled to 12 frames to reduce computational cost and give equal weight to episodes of different lengths. We retain the first frame, last frame, and ten evenly spaced middle frames. All metrics are reported as mean $\pm$ standard deviation over five seeds. For attribution, the evaluation target is the \emph{empirical per-corruption attribution},
\begin{equation}
    \bar{\Delta}_k(c)=\hat P(y=1\mid c)-\hat P(y=1\mid c^{\setminus k}),
\end{equation}
where $\hat P(y=1\mid c)$ is the empirical failure rate over test episodes under configuration $c$. For hardware, empirical attribution targets use all collected episodes, while predictions are evaluated on held-out episodes. We clip the resulting difference at zero, consistent with \cref{eq:estrf}. In simulation, rates are pooled across all 10 tasks and initial states. The resulting $\bar{\Delta}_k$ serves as the
configuration-level evaluation target for attribution, but does not cover variation in attribution across individual scenes.

We evaluate attribution on each held-out episode's initial observation. We report attribution \textit{mean absolute error} MAE, \(|\hat{\Delta}_k-\bar{\Delta}_k(c)|\), over present corruptions, separately for single- and multi-corruption episodes. For multi-corruption failures, \textit{regret} is reported, \(\bar{\Delta}_{k^\ast}(c)-\bar{\Delta}_{\hat{k}}(c)\), for \(\hat{k}=\arg\max_k\hat{\Delta}_k\), and measures the cost of selecting \(\hat{k}\) instead of the optimal \(k^\ast\). The predicted ranking used for regret is computed from pre-ReLU risk differences. Presence accuracy is averaged over corruptions at \(\hat{\rho}_k>0.5\). VLM baselines are evaluated on the same observations, with their \(0\)--\(10\) responsibility scores rescaled to \([0,1]\) as \(\hat{\Delta}\).

We also evaluate single-frame failure prediction at different episode positions using ROC-AUC between predicted risk $\hat r$ and the episode outcome $y$. At each position, nominal and corrupted test episodes are evaluated separately. ROC-AUC is computed for each training seed and reported as mean $\pm$ standard deviation.
\subsubsection{Training Details}
We use the AdamW~\cite{loshchilov2019adamw} optimizer with learning rate $2\times10^{-3}$ and weight decay $10^{-4}$. The failure head also has a small $L_2$ penalty ($10^{-3}$) on its weights, which regularizes the head early in training. We train for $200$ epochs with batch size $256$ over $5$ seeds. The attributor is a two-latent MLP over $1161$-d SigLIP image-and-text features (1152-d image features concatenated with a $9$-d PCA projection of the task embedding), with a $32$-d scene latent $z_s$, a corruption latent $z_c$ of $K$ four-dimensional blocks, and presence/failure/adversary heads of hidden width $128$. The loss-term weights are $\lambda_{\mathrm{pres}}=1$, $\lambda_{\mathrm{anch}}=0.1$, $\lambda_{\mathrm{cons}}=3$, and $\lambda_{\mathrm{adv}}$ ramped linearly from $0$ to $20$ over the first $40$ epochs. The loss weights balance the prediction and representation objectives. We selected them through individual sweeps using OpenVLA validation metrics and used the same weights for both simulation policies and the hardware experiments.

\subsubsection{Baselines}
We compare against a monolithic, non-factorized baseline, \baselinename{Monolithic}, that first predicts corruption logits from SigLIP features $x_t$ and subsequently predicts failure using $x_t$ and the corruption logits. We also evaluate few-shot attribution with \baselinename{VLM (Terra)}, \baselinename{VLM (Sol)}, and \baselinename{VLM (Opus\,5)} (GPT-5.6 Terra, GPT-5.6 Sol, and Claude Opus 5, respectively). Each VLM receives the query frame, one successful and one failed example per corruption, and a nominal pair, then identifies present corruptions and assigns each a \(0\)–\(10\) responsibility score. Finally, \baselinename{Ours (SAFE)} replaces the SigLIP input of our full method, \baselinename{Ours (SigLIP)}, with internal policy features extracted following~\cite{gu2025safe}, holding the remaining architecture and training procedure fixed.

We include an \baselinename{Oracle} reference using ground-truth corruption labels and empirical failure rates for each corruption combination. It computes attributions as the clipped risk reduction from removing each corruption. Since these rates also define the empirical attribution targets, it achieves zero MAE and regret by construction. The oracle reveals whether our model captures execution context and corruption intensity beyond combination-level average failure rates.

\section{Results}\label{sec:results}
Below we discuss the key results from our experimental analysis using several research questions.
\begin{table}[t]
    \centering
    \caption{Loss ablation (OpenVLA, \baselinename{Ours (SigLIP)}; five-seed mean $\pm$ standard deviation). Pres.: presence accuracy (\%). MAE: single-corruption attribution MAE. Reg.: regret on multi-corruption failure episodes. $\|z_c^{(k)}\|_{\varnothing}$: mean norm of block $k$ when corruption $k$ is absent ($0$ ideal). $z_s$\,leak: AUC (\%) of predicting presence from $z_s$ ($50$ ideal, $100$ full leakage). Ablations use separately trained models.}
    \label{tab:ablation}
    \small
    \setlength{\tabcolsep}{2.5pt}
    \renewcommand{\arraystretch}{1.15}
    \resizebox{\columnwidth}{!}{%
    \begin{tabular}{lccccc}
        \toprule
        Variant & Pres. & MAE\,$\downarrow$ & Reg.\,$\downarrow$ & $\|z_c^{(k)}\|_{\varnothing}$\,$\downarrow$ & $z_s$\,leak \\
        \midrule
        \textbf{Full}          & $99.4$ & $0.092_{\pm.017}$ & $0.010_{\pm.004}$ & $0.31_{\pm.07}$ & $75.1_{\pm8.2}$ \\
        \midrule
        $-$adversary           & $99.5$ & $0.080_{\pm.010}$ & $0.014_{\pm.003}$ & $0.21_{\pm.04}$ & $99.8_{\pm0.1}$ \\
        $-$anchor              & $99.4$ & $0.089_{\pm.021}$ & $0.040_{\pm.031}$ & $30.3_{\pm3.1}$ & $78.0_{\pm9.0}$ \\
        $-$consistency         & $99.5$ & $0.132_{\pm.015}$ & $0.016_{\pm.011}$ & $0.06_{\pm.04}$ & $72.1_{\pm6.9}$ \\
        $-$INLP                & $99.5$ & $0.084_{\pm.011}$ & $0.016_{\pm.005}$ & $0.35_{\pm.03}$ & $82.6_{\pm10.7}$ \\
        $-$presence            & $49.3$ & $0.370_{\pm.033}$ & $0.245_{\pm.061}$ & $0.52_{\pm.14}$ & $68.6_{\pm6.8}$ \\
        \midrule
        $-$INLP$-$adv          & $99.5$ & $0.080_{\pm.011}$ & $0.012_{\pm.004}$ & $0.22_{\pm.01}$ & $99.9_{\pm0.1}$ \\
        $-$INLP$-$adv$-$anchor  & $99.4$ & $0.084_{\pm.018}$ & $0.015_{\pm.007}$ & $22.9_{\pm4.0}$ & $99.9_{\pm0.1}$ \\
        \bottomrule
    \end{tabular}%
    }

\end{table}
\subsection{Is corruption detection sufficient for failure attribution?}
\label{sec:detectionattribution}
We first ask whether accurate corruption detection is sufficient for failure attribution. Corruption presence is easy to predict; our model achieves \(99.4\%\) and \(99.3\%\) accuracy on OpenVLA and \(\pi_{0.5}\) in simulation, respectively, while the monolithic and stronger VLM baselines exceed \(90\%\). However, accurate detection does not ensure accurate estimates of a corruption's effect on policy performance.

As shown in \cref{tab:main_attribution}, our method achieves lower single- and multi-corruption attribution MAE than the monolithic and VLM baselines. \baselinename{Ours (SigLIP)} obtains single-/multi-corruption MAE of $0.095/0.101$ for OpenVLA and $0.019/0.064$ for $\pi_{0.5}$.

Despite high presence accuracy, the VLMs' policy-agnostic scores yield substantially higher MAE and regret, showing that detecting degradation is not the same as estimating its policy-specific effect.
\cref{fig:calibration_grid} visualizes the difference between predicted and empirical per-corruption attribution across corruption conditions: our per-condition estimates track the empirical attributions closely (MAE $0.09$ and $0.05$ for OpenVLA and $\pi_{0.5}$), while the VLM's scores are poorly calibrated with them (MAE $0.30$ and $0.32$).

\subsection{Does our proposed attribution scheme capture co-occurring corruptions and their interactions?}
\label{sec:cooccurrence}
\cref{tab:main_attribution} shows that our method remains accurate under co-occurring corruptions, with multi-corruption MAE remaining low (\(0.101\) for OpenVLA and \(0.064\) for \(\pi_{0.5}\)). It also achieves the lowest regret on both policies: \(0.011\) versus \(0.098\) for the strongest VLM on OpenVLA, and \(0.005\) versus \(0.026\) on \(\pi_{0.5}\). Thus, our selected corrections have a small average loss in removal benefit.

Our method also captures interactions because the failure head jointly models the full corruption latent \(z_c\). As \cref{fig:interaction} shows, OpenVLA exhibits strong non-additivity when either corruption alone can cause failure. For light+view, removing either alone has little effect (light: \(0.04\), view: \(0.19\)), whereas removing both reduces risk by \(0.86\), close to the empirical \(0.85\). Conversely, pairs involving the largely inert object-add show no compounding effect. In contrast, \(\pi_{0.5}\), which is primarily sensitive to blur, exhibits mostly additive interactions. Thus, the model recovers policy-specific interaction structure rather than imposing it.

\subsection{Does the model capture variations in failure risk and attribution beyond corruption presence?}
\label{sec:beyondpresence}
We evaluate whether the model captures execution context
and corruption intensity beyond corruption presence alone.
Because the \baselinename{Oracle} assigns identical predictions to all observations of a given corruption combination, any per-episode or per-intensity variation our model recovers reflects information beyond configuration-level rates.

First, \cref{tab:failrecog} reports failure-prediction ROC-AUC at different episode positions within held-out OpenVLA episodes. On nominal episodes the \baselinename{Oracle} is at chance ($0.500$), while our model rises from $0.510$ at the initial frame to $0.723$ at the final frame; on corrupted episodes it ranges from $0.917$ to $0.932$, exceeding the \baselinename{Oracle}'s $0.862$ everywhere. Discrimination strengthens later in the episode, as the image gives clearer evidence of the outcome.

\begin{figure}[t]
    \centering
    \includegraphics[width=\linewidth]
        {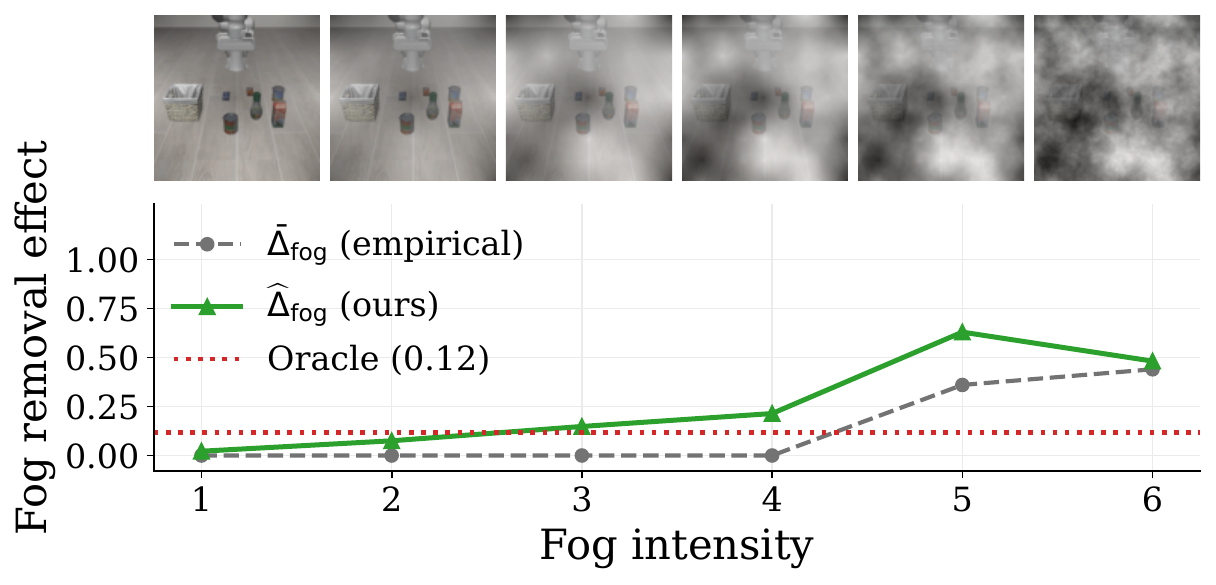}

    \caption{
        Fog attribution across six intensities for OpenVLA with \baselinename{Ours (SigLIP)}. Each intensity is applied to the same held-out scenes. Level 4 occurs in the training data, while the others are unseen. Our predictions respond to intensity, while the oracle retains its original test-set lookup value across the sweep.
    }
    \label{fig:latent_severity}
\end{figure}

For intensity, we evaluate the same trained model on six ordered fog levels (five unseen in training, spanning within and beyond the training range) applied to matched held-out scenes. Since fog presence is unchanged across the sweep, the \baselinename{Oracle} is constant, whereas our fog attribution varies with intensity (\cref{fig:latent_severity}), capturing variation the \baselinename{Oracle} cannot express. Together, these show the model conditions on execution context and intensity, not just corruption presence.

\subsection{How do various components enable accurate counterfactuals?}
\label{sec:latentfact}

The \baselinename{Monolithic} baseline in \cref{tab:main_attribution} illustrates the bypass problem motivating our factorization: the failure head can use corruption information already present in SigLIP features without relying on the explicit corruption logits, eliminating the ability for counterfactual intervention. Our model instead encourages corruption information to pass through \(z_c\). Ablations on the loss terms on OpenVLA (\cref{tab:ablation}) show that removing presence loss ($\mathcal{L}_{\mathrm{pres}}$) reduces detection from \(99.4\%\) to \(49.3\%\) and raises single-corruption MAE to \(0.370\). Removing counterfactual consistency loss ($\mathcal{L}_{\mathrm{cons}}$) increases MAE (\(0.092\) to \(0.132\)) and increases regret (\(0.010\) to \(0.016\)), confirming its role in aligning masked predictions with corruption-absent observations. Removing the anchor loss ($\mathcal{L}_{\mathrm{anch}}$) leaves MAE largely unchanged ($0.092$ to $0.089$) but sharply increases the absent-block norm from \(0.31\) to \(30.3\) and regret to \(0.040\), showing that zero no longer represents absence reliably.

The adversarial loss ($\mathcal{L}_{\mathrm{adv}}$) and INLP primarily prevent corruption leakage through \(z_s\). Removing either increases probe AUC from \(75.1\) to \(99.8\) or \(82.6\), respectively, with limited effect on MAE. Thus, consistency and anchoring ensure accurate counterfactuals, while adversarial training and INLP preserve latent disentanglement.

\subsection{Are input representations or internal policy features more effective for attribution?}\label{sec:safe}
We compare SigLIP image-and-text features with internal policy features extracted following SAFE~\cite{gu2025safe}, using the same attribution framework. SigLIP matches or improves on internal features across the board (\cref{tab:main_attribution}). For OpenVLA, SigLIP yields lower single-corruption MAE ($0.095$ versus $0.104$), lower multi-corruption MAE ($0.101$ versus $0.117$), and lower regret ($0.011$ versus $0.029$). The same holds for $\pi_{0.5}$: $0.019/0.064$ MAE with SigLIP versus $0.048/0.075$ with internal features, and lower regret ($0.005$ versus $0.028$).

SigLIP yields the lowest error across both policies without requiring access to policy internals, enabling attribution for black-box policies. Its shared feature space also supports novel-corruption detection. We fit class-conditional Gaussians with shared covariance over the SigLIP features and score each frame by its Mahalanobis distance~\cite{lee2018mahalanobis} to the nearest class. In a leave-one-corruption-out evaluation, this separates the held-out novel corruption from in-distribution frames with mean AUROC $0.88$.

\subsection{Does the method transfer to hardware and online monitoring?}
We evaluate on the DROID hardware platform~\cite{khazatsky2024droid} using the pretrained \(\pi_{0.5}\)-DROID checkpoint and an attribution model trained only on hardware episodes. As shown in \cref{tab:hardware}, predicted attributions are $0.96$, $0.56$, $0.27$, and $0.03$ for light, fog, blur, and object-add, respectively, compared with empirical attributions of $0.96$, $0.56$, $0.24$, and $0.04$. The model identifies near-darkness as most damaging and object-add as nearly harmless. The VLM assigns object-add a substantially larger score ($0.35$).
The model supports online monitoring: \(\hat r\) signals early-stopping, while \(\hat\Delta_k\) identifies the responsible corruptions. In \cref{fig:hardware_rollout}, \(\hat r\) remains low for a distractor, increases with blur and fog, and approaches \(1.00\) for near-darkness.
Thus, the model tracks corruption contributions online and can halt execution before failure.

\begin{table}[t]
    \centering
    \caption{
        Hardware attribution ($\pi_{0.5}$-DROID, ``put the cube in the bowl''; five-seed mean $\pm$ standard deviation). Empirical per-corruption attribution $\bar\Delta_k$ (all episodes), our estimate $\hat\Delta_k$ (held-out), and \baselinename{VLM (Sol)}'s responsibility score rescaled to $[0,1]$.
    }
    \label{tab:hardware}
    \small
    \setlength{\tabcolsep}{5pt}
    \renewcommand{\arraystretch}{1.1}
    \begin{tabular}{lccc}
        \toprule
        Corruption & $\bar\Delta_k$ (emp.) & $\hat\Delta_k$ (ours) & VLM score \\
        \midrule
        Light        & $0.96$ & $0.96_{\pm.01}$ & $1.00$ \\
        Fog          & $0.56$ & $0.56_{\pm.05}$ & $0.70$ \\
        Blur         & $0.24$ & $0.27_{\pm.03}$ & $0.45$ \\
        Object       & $0.04$ & $0.03_{\pm.01}$ & $0.35$ \\
        \bottomrule
    \end{tabular}
\end{table}

\section{Discussion and Conclusion}
In this work, we introduced a counterfactual method for attributing VLA failures to visual corruptions. By disentangling scene and corruption latents and jointly predicting failure from both, we estimate each corruption's attribution, \(\hat\Delta_k\), from a single observation without clean-image pairs or additional rollouts. Experiments with OpenVLA and \(\pi_{0.5}\) in LIBERO and \(\pi_{0.5}\)-DROID on hardware show that detection alone is insufficient for attribution: VLM baselines rely on policy-agnostic visual priors, whereas our estimates recover policy-specific attribution rankings, capture interactions among co-occurring corruptions, and support online hardware monitoring.

Our method is limited to known corruptions with training-time presence labels. Future work could extend attribution to unseen corruptions and collect larger datasets for task- and state-specific analysis. Integrating attribution with automated recovery is another promising direction for preventing deployment failures.

\bibliographystyle{IEEEtran}
\bibliography{root}

\end{document}